%% file: main_arxiv.tex
\documentclass{article}

\usepackage{microtype}
\usepackage{graphicx}
\usepackage{subfigure}
\usepackage{booktabs} % for professional tables

\usepackage{hyperref}

\usepackage[accepted]{icml2024}

\usepackage{amsmath}
\usepackage{amssymb}
\usepackage{mathtools}
\usepackage{amsthm}

\usepackage[capitalize,noabbrev]{cleveref}

\theoremstyle{plain}

\theoremstyle{definition}

\theoremstyle{remark}

\usepackage[textsize=tiny]{todonotes}

\icmltitlerunning{\ours: An efficient AI weather forecasting model at 1.5º resolution}

\newcommand{\ours}{\begin{sc}ArchesWeather\end{sc}}
\newcommand{\sct}[1]{\begin{sc}#1\end{sc}}
\newcommand{\bfu}[1]{\textbf{\underline{#1}}}

\begin{document}

\twocolumn[
\icmltitle{\ours: An efficient AI weather forecasting model at 1.5º resolution}

% It is OKAY to include author information, even for blind
% submissions: the style file will automatically remove it for you
% unless you've provided the [accepted] option to the icml2024
% package.

% List of affiliations: The first argument should be a (short)
% identifier you will use later to specify author affiliations
% Academic affiliations should list Department, University, City, Region, Country
% Industry affiliations should list Company, City, Region, Country

% You can specify symbols, otherwise they are numbered in order.
% Ideally, you should not use this facility. Affiliations will be numbered
% in order of appearance and this is the preferred way.
\icmlsetsymbol{equal}{*}

\begin{icmlauthorlist}
\icmlauthor{Guillaume Couairon}{inria}
\icmlauthor{Christian Lessig}{ecmwf}
\icmlauthor{Anastase Alexandre Charantonis}{inria,ensiie}
\icmlauthor{Claire Monteleoni}{inria,boulder}
\end{icmlauthorlist}

\icmlaffiliation{inria}{Inria, France}
\icmlaffiliation{ensiie}{ENSIEE, France}
\icmlaffiliation{ecmwf}{ECMWF, Germany}

\icmlaffiliation{boulder}{University of Colorado Boulder, USA}

\begin{small}
\vskip 0.3em
    \qquad \qquad \qquad \qquad 
    $^1$Inria, France \quad $^2$ENSIIE, France \quad $^3$ECMWF, Germany \quad $^4$University of Colorado Boulder, USA
\end{small}

%\icmlcorrespondingauthor{Guillaume Couairon}{guillaume.couairon@gmail.com}

% You may provide any keywords that you
% find helpful for describing your paper; these are used to populate
% the "keywords" metadata in the PDF but will not be shown in the document

\icmlkeywords{Machine Learning, AI weather forecasting}
\vskip 0.3in
]

%\renewcommand{\ICML@appearing}{}
% this must go after the closing bracket ] following \twocolumn[ ...

% This command actually creates the footnote in the first column
% listing the affiliations and the copyright notice.
% The command takes one argument, which is text to display at the start of the footnote.
% The \icmlEqualContribution command is standard text for equal contribution.
% Remove it (just {}) if you do not need this facility.

%\printAffiliationsAndNotice{}  % leave blank if no need to mention equal contribution
%\printAffiliationsAndNotice{}%\icmlEqualContribution} % otherwise use the standard text.

\input{core}

\end{document}

%% file: core.tex
\begin{abstract}
One of the guiding principles for designing AI-based weather forecasting systems is to embed physical constraints as inductive priors in the neural network architecture. A popular prior is locality, where the atmospheric data is processed with local neural interactions, like 3D convolutions or 3D local attention windows as in Pangu-Weather. On the other hand, some works have shown great success in weather forecasting without this locality principle, at the cost of a much higher parameter count.

In this paper, we show that the 3D local processing in Pangu-Weather is computationally sub-optimal. We design \ours, a transformer model that combines 2D attention with a column-wise attention-based feature interaction module, and demonstrate that this design improves forecasting skill.

\ours~is trained at 1.5º resolution and 24h lead time, with a training budget of a few GPU-days and a lower inference cost than competing methods. An ensemble of four of our models shows better RMSE scores than the IFS HRES and is competitive with the 1.4º 50-members NeuralGCM ensemble for one to three days ahead forecasting.

Our code and models are publicly available at \url{https://github.com/gcouairon/ArchesWeather}.

\end{abstract}

\section{Introduction}

\begin{figure}[t]
    \centering
    \includegraphics[width=1.1\linewidth]{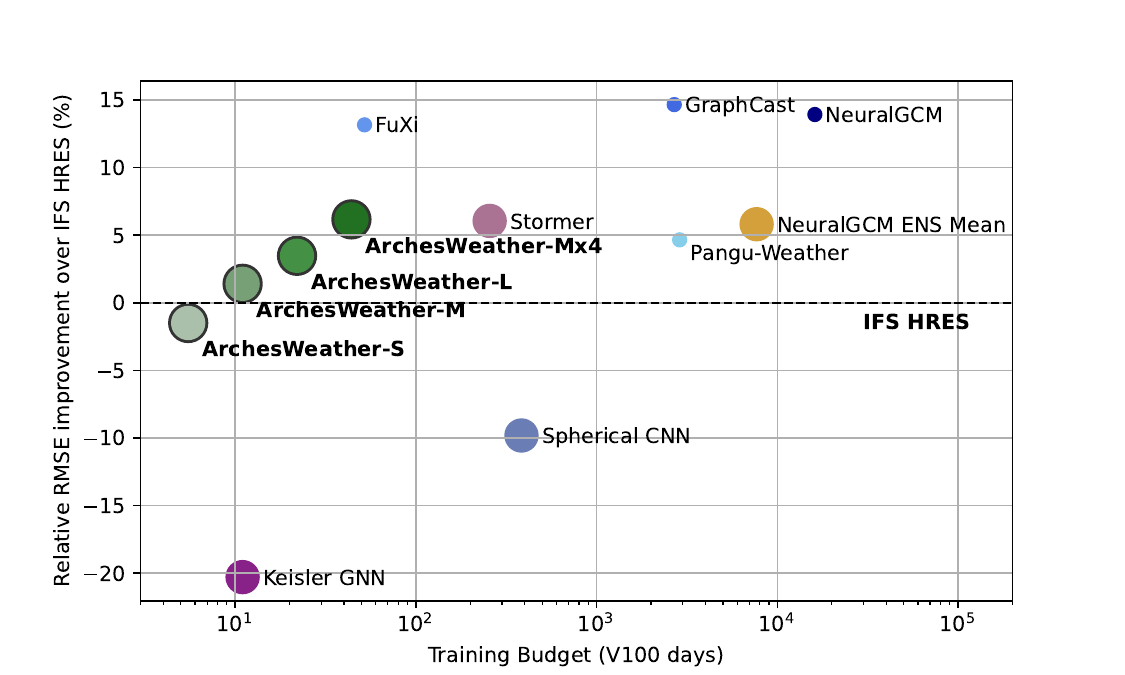}
    \vspace{-1.5em}
    \caption{Relative RMSE improvement over the IFS HRES as a function of training computational budget, averaged for key upper air variables (Z500, Q700, T850, U850 and V850) and lead times of 24h/48h/72h. Circle size indicate training resolution: small circles for 0.25º/0.7º, big circles for 1º/1.4º/1.5º. \ours~reaches competitive forecasting performance with a much smaller training budget.}
    \vspace{-1em}
    \label{fig:teaser}
\end{figure}

The field of weather forecasting is undergoing a revolution, as AI models trained on the ERA5 reanalysis dataset \cite{hersbach2020era5} can now outperform IFS-HRES, the reference numerical weather prediction model developed by the European Center for Medium-Range Weather Forecasting (ECMWF), with inference costs that are orders of magnitude lower \cite{bi2022pangu, pathak2022fourcastnet, lam2022graphcast, chen2023fuxi, nguyen2023scaling, guo2024fourcastnext, kochkov2023neural}.The neural network architectures of these models are adopted from the computer vision community, usually by adding priors related to the specificity of processing physical fields on a 3D spherical atmosphere (local 3D attention for Pangu-Weather; Fourier Spherical Operators for FourCastNet; Graph Neural Networks on a spherical mesh for GraphCast; Dynamical core for NeuralGCM). Adding these physical priors usually served two goals: (i) AI models that have more priors are more interpretable since they more closely relate to their numerical counterparts, which increases trust in these models; (ii) networks with more physical priors generalize better and can reach the same accuracy with less parameters and memory footprint.

However, recent works have started questioning this second assumption, showing that architectures with less physical priors can also generalize well with a smaller training cost \cite{nguyen2023scaling, chen2023fuxi, lessig2023atmorep}, which might hint that models with more physical priors and less parameters are harder to train. These works have adapted vision transformers \cite{dosovitskiy2020image} by considering ERA5 as latitude/longitude images, and concatenating upper-air weather variables in the channel dimension. This concatenation requires a lot more parameters than 3D processing, so these works still rely on very large neural networks (300M parameters for Stormer, 1.5B for FuXi).

In this paper, we identify a limitation of 3D local attention, used in the Pangu-Weather architecture. Inside the network, only the features for neighboring pressure levels interact, mimicking the physical principle that air masses only interact locally at short timescales. We find that despite its connection with physics, this prior is computationally sub-optimal and we design a global \textbf{C}ross-\textbf{L}evel \textbf{A}ttention-based interaction layer (dubbed CLA) to overcome this limitation.

We also show that there is a small distribution shift in ERA5 before and after 2000, which we attribute to shifts in the observation system, and we improve forecasting by fine-tuning on recent data.

Our model, dubbed \ours, is trained at 1.5º resolution and 24h lead time, in three versions: S (49M params), M (89M params), L (164M params). Our M version reaches competitive RMSE scores with a computational budget orders of magnitude less than competing architectures (see Figure \ref{fig:teaser}). An ensemble average of four M models is competitive with the 1.4º NeuralGCM ensemble with 50 members \cite{kochkov2023neural} for a lead time of one to three days. Our work paves the way for training weather models at 1.5º on academic resources, only requires to download less than 1 TB of data and provides cheap inference: a single 24h forecast with the M model takes $\sim$0.25s on a A100 GPU card.

In summary, our contributions are:
\vskip -0.4em
\begin{itemize}
    \item We show that the 3D local attention in Pangu-Weather is computationally sub-optimal and we design a non local cross-level attention layer that boosts performance.
    \vskip -0.1em
    \item We show that some additional benefit can be gained by fine-tuning our model on recent ERA5 data.
    \vskip -0.1em
    \item \ours~is competitive with the state-of-the-art while requiring only a few GPU-days to train and can be run cheaply at 1.5º.
    \vskip -0.1em
\end{itemize}

\section{Methods}

We tackle the task of AI-based weather forecasting, and denote with $(X_t)$ the historical trajectory of weather variables. We optimize a neural network to predict the next state $X_{t+\delta t}$ given an input state $X_t$, where $\delta t$ is called the \textit{lead time}, which is set to 24h for the remainder of the paper.

\subsection{Data, Evaluation and Metrics}

We train our models on the ERA5 dataset regridded to 1.5º resolution, which is the standard used for evaluation at the World Meteorological Organization (WMO). We use 6 upper air variables (temperature, geopotential, specific humidity, wind components U, V and W) at 13 pressure levels, and 4 surface variables (2m temperature, mean sea-level pressure, 10m wind U and V), sampled every 6h.

Following the standard in Weatherbench 2 \cite{rasp2023weatherbench}, we train on the ERA5 data from 1979 to 2018, validate on the year 2019, and test our models at 00/12UTC for each day of 2020. Models are evaluated with the latitude-weighted Root Mean Square Error (RMSE). We also define a metric called RRH (average Relative RMSE improvement over the IFS HRES), to get a representative score across key weather variables, detailed in Appendix \ref{app:metrics}.

\subsection{Architecture}

Our neural network architecture is a 3D Swin U-Net transformer  \cite{liu2021swin, liu2022swin} with the Earth-specific positional bias, largely inspired by the Pangu-Weather architecture. The surface and upper-air variables are first embedded into a single tensor of size $(d, Z, H, W)$ where $d$ is the embedding dimension, $Z$ the vertical dimension, $H$ and $W$ the latitude and longitude dimensions. this tensor is then processed by the U-Net transformer, and is projected back to surface and upper-air at the end. The standard is to use a strided deconvolution layer for this final projection \cite{bi2022pangu, chen2023fuxi}, however it tends to produce un-physical artefacts (see Figure \ref{fig:convhead} in Appendix). Instead, we use a deconvolution head with bilinear upsampling followed by a standard convolution (see Appendix \ref{app:convhead}). Finally, following GraphCast, we provide additional information to the model (hour and month of desired forecast) with adaptive Layer Normalization \cite{perez2018film}.

\subsection{Improving efficiency with Cross-Level Attention (CLA)}

\begin{figure}[ht]
    \centering
    \includegraphics[width=\linewidth]{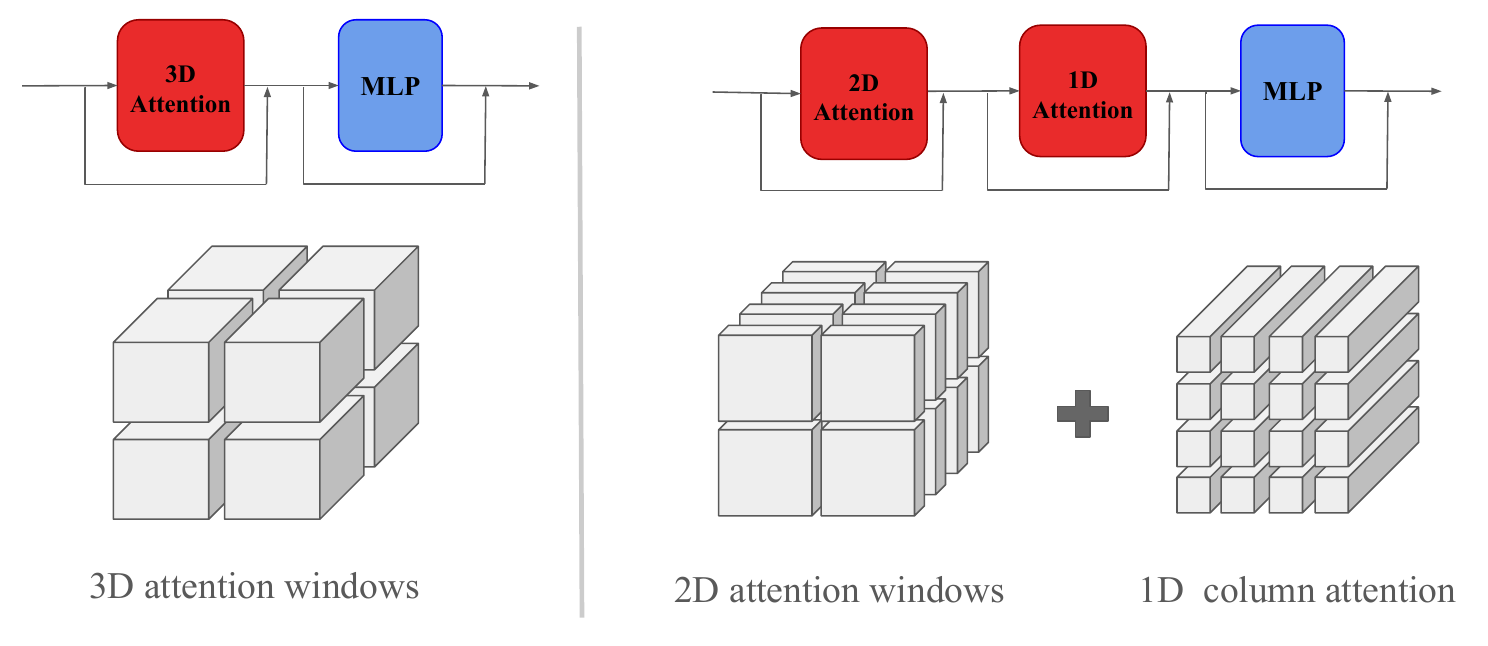}
    \vspace{-1.8em}
    \caption{Comparison of attention schemes used in Pangu-Weather (left) versus ours (right).}
    \label{fig:attention}
    \vspace{-0.4em}
\end{figure}

The attention scheme in a Swin layer \cite{liu2021swin} consists in splitting the input tensor in non-overlapping windows, where a self-attention layer processes each window independently. Then, data is shifted by half a window to compute the next self-attention layer, allowing interaction between the different attention windows. In Pangu-Weather, the input tensors are split in 3-dimensional windows of size $(2, 6, 12)$: hence, along the vertical $Z$ dimension, only the features for neighboring pressure levels interact, mimicking the physical principle that air masses only interact locally at short timescales. This inductive prior is meant to have the neural network roughly reproduce physical interaction phenomena and reduce the number of parameters needed.

\paragraph{Limitation.} From a computational perspective, this prior is a limitation since computations for similar phenomena happening at different atmospheric layers are performed independently in parallel. Global vertical interaction would allow sharing such computations, allocating resources more efficiently. Computations for complex variables can also be spread across levels faster, to reach lower error. Finally, from a physical perspective, having vertical interaction can allow to detect the vertical profile of the atmosphere and to adjust processing accordingly.

Before presenting our proposed solution, we mention two other potential methods and their caveats. First, one could increase the attention window size, e.g. to $(4, 6, 12)$ instead of $(2, 6, 12)$, to accelerate exchange of information along the vertical dimension, but this decreases inference speed due to the quadratic cost of attention in the sequence length. Second, some works use a more standard 2D transformer \cite{nguyen2023scaling, chen2023fuxi} and stack variables across pressure levels in a single vector at each spatial position. This comes at the cost of an increased parameter count: With $Z$ pressure levels (after embedding), the linear and attention layer need $O(d^2Z^2)$ parameters, with $d$ being the embedding dimension for a single pressure level. As a result, Stormer uses a ViT-L with 300M parameters, and FuXi uses a SwinV2 architecture with 1.5B parameters.

\begin{figure}[ht]
    \centering
    \includegraphics[width=\linewidth]{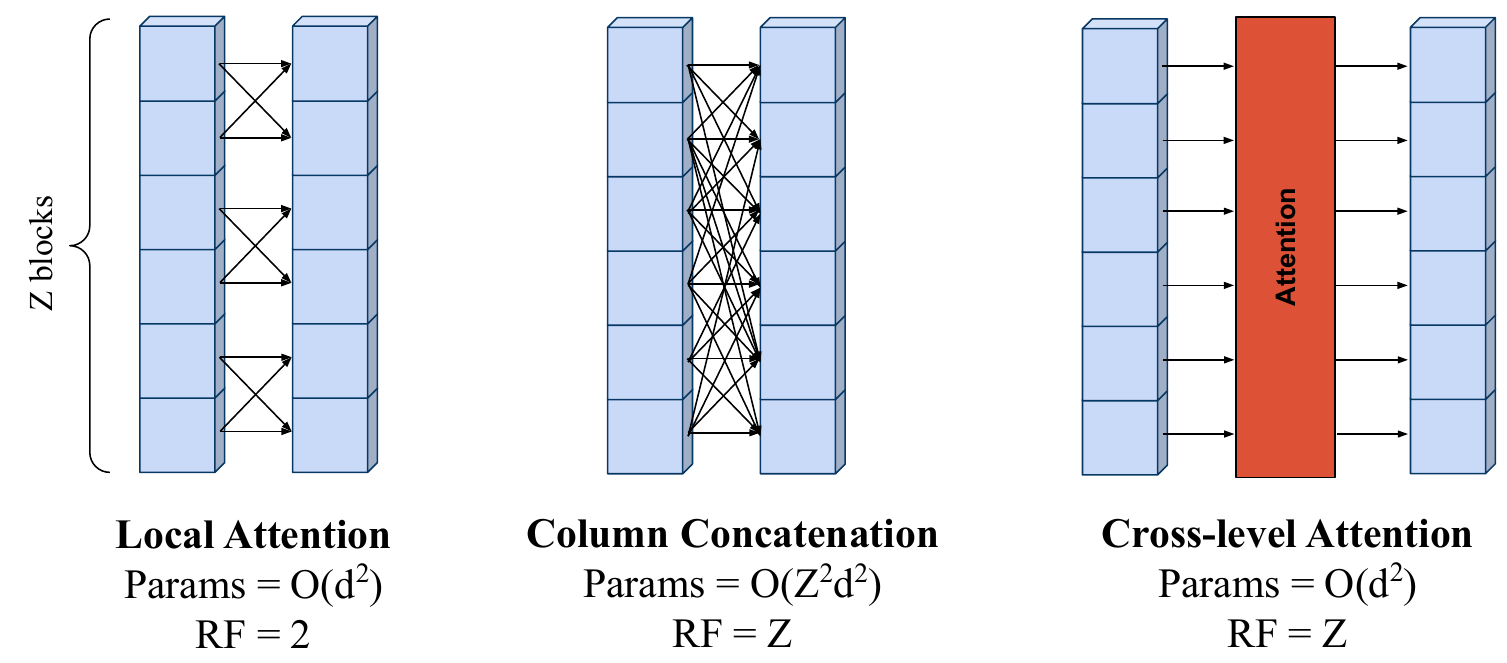}
    \vspace{-1.8em}
    \caption{Comparison of attention schemes used in Pangu (Local Attention, left), Stormer/FuXi (Concatenated columns, middle) and ours (Cross-level Attention, right). For each scheme, a single vertical column is represented to illustrate how each layer processes column-wise information. RF stands for Receptive Field.}
    \label{fig:colatt}
    \vskip -0.5em
\end{figure}

\paragraph{Proposed solution. } We propose to make all vertical features interact by adding a column-wise attention mechanism dubbed Cross-Level Attention (CLA), that processes data along the vertical dimension of the tensor only. By considering column data as a sequence of size $Z$, the number of parameters in this attention module is $O(d^2)$ and does not depend on $Z$. 
We also remove the vertical interaction from the original implementation by using horizontal attention windows of shape $(1, 6, 12)$, which reduces the attention cost. The resulting attention scheme shares similarities with axial attention \cite{ho2019axial} with a decomposition of attention in two parts: column-wise attention and local horizontal 2D attention. See Figures \ref{fig:attention} and \ref{fig:colatt} for an illustration of our proposed attention scheme, compared to other attention methods. Axial attention has also been used in MetNet-3 \cite{andrychowicz2023deep} and SEEDS \cite{li2023seeds}.

%\subsection{Other architectural choices}

%In Pangu-Weather, a lot of parameters come from the earth positional bias, which depend on the input resolution. At 1.5 degree resolution, the Pangu-Weather has only 24M parameters, compared to 256M for the original model. Since we expect the task of weather prediction at 1.5º to be more complex than at 0.25º, we use a patch size of 2 instead of 4, and double the number of layers, resulting in a base model with 88M parameters. 

%We also identify two limitations of Pangu-Weather that we address: first, the skip connection was only concatenated before the deconvolution layer at the end, leaving no possibility for complex interaction between the features from the skip connection and the features from the deeper, coarser resolution branch. We address this by concatenating the skip connection earlier in the architecture.

\input{figs/main_table}

\subsection{Training details}

Our model comes in three versions: \sct{ArchesWeather-S}, 16 transformer layers (49M parameters); \sct{ArchesWeather-M}, 32 layers (89M parameters); \sct{ArchesWeather-L}, 64 layers (164M parameters). We train all models for $320k$ steps; training the M model takes around 2.5 days on 2 A100 GPUs. More details can be found in Appendix \ref{app:training-details}.

\begin{figure}[ht]
    \centering
    \includegraphics[width=\linewidth]{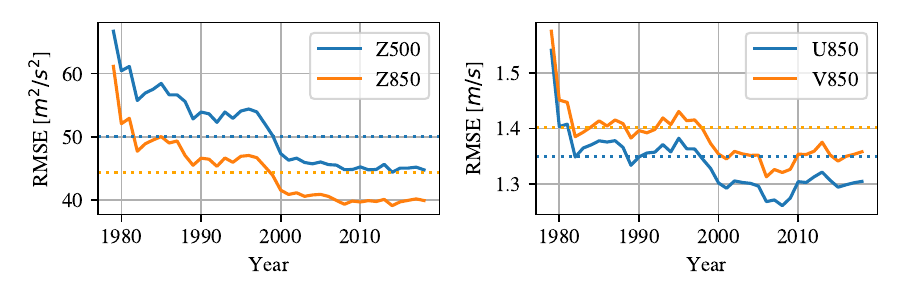}
    \vspace{-1.8em}
    \caption{Geopotential (left) and wind speed (right) RMSE of a model w/o fine-tuning, for each year in the training set. Test RMSE (year 2020) are shown in dotted lines.}
    \label{fig:finetunefig}
    \vskip -0.2em
\end{figure}

Next, we find that forecasting models have a larger error in the first half of the training period 1979-2018 (see Figure \ref{fig:finetunefig}, which we attribute to ERA5 being less constrained in the past due to a lack of observation data. To overcome this distribution shift, we use recent ERA5 data (2007-2018) for fine-tuning our models from steps $250k$ to $300k$.

As commonly done in similar works \cite{lam2022graphcast, chen2023fuxi, nguyen2023scaling}, we fine-tune our models for 20k steps on auto-regressive rollouts of length $2$ to $4$, see details in appendix \ref{app:multistep}.

Finally, we train small ensembles of our models, by independently training multiple models with different random seeds, and then averaging their outputs at inference time. We call these models \sct{ArchesWeather-Mx4} (four M models) and \sct{ArchesWeather-Lx2} (two L models).

\vskip 1.5em
\section{Experiments}

\subsection{Main results}

Table \ref{table:main} shows RMSE scores of \ours~compared to state-of-the-art ML 
weather models, including Pangu-Weather and GraphCast, SphericalCNN \cite{esteves2023scaling}, NeuralGCM at 1.4º (50 members ensemble), and Stormer.
Data is from WeatherBench 2, except Stormer where we evaluated outputs provided by the authors.

The \sct{ArchesWeather-M} base model largely surpasses the SphericalCNN model for upper-air variables, with a training budget of around 10 V100-days, 40 times smaller. At 24h lead time, the \ours~ensemble version outperforms the 1.4º NeuralGCM ensemble (50 members) on upper-air variables. They perform on par with the original Pangu-Weather(0.25º) and Stormer(1.4º), except for wind variables (U850, V850, U10, V10) where notably Stormer is consistently better. This might be due to the higher training budget (256 V100-days), bigger models, or averaging outputs from 16 model forward passes (more details in Appendix \ref{app:sota}). Investigating this discrepancy is left for future work. Interestingly, this 24h advantage for wind variables disappears at longer lead times, see Appendix \ref{app:multistep}.
Finally, our model shows very good RMSE scores at longer lead times, as shown in Appendix \ref{app:multistep}.

\vskip -0.8em
\subsection{Ablation}
\vskip -0.5em

\input{figs/ablation_table}

Our main ablation experiment is presented in Table \ref{abl}, where we compare models without multi-step fine-tuning.
For a fair comparison between models, we decrease the embedding dimension (by about 5\%) when using CLA, so that all types of models have roughly the same parameter count. Compared to Pangu-Weather (retrained in the same setting as us), our model without Cross-Level Attention or fine-tuning largely improves performance (rows D compared to rows A), which is largely due to the methodology improvements from GraphCast (predicting $X_{t+\delta t} - X_t$ instead of $X_t$, including the wind vertical component, conditioning on the day and month) and the convolutional head.
Adding on top our proposed Cross-Level Attention scheme significantly improves scores (Rows C versus D), reducing by half the RMSE difference with the IFS HRES. \ours~with 16 layers reaches lower error than using 32 layers without CLA (e.g. Z500 RMSE of 50.6 vs 51.8).
Finally, finetuning the model on recent data only for the last 50k steps brings some small additional benefit (rows C versus B).

\section{Conclusion}

We have presented \ours, a weather model that operates at 1.5º, only requires a few GPU-days to train with a reasonably sized dataset ($<$ 1TB), yet reaches similar performance as some models trained with a much higher computational budget. We also find that fine-tuning on recent data slightly improves skill.
\ours is however less suited for applications that require a better resolution, like cyclone tracking, or regional forecasting. The outputs of our model could potentially be downscaled to a finer resolution and projected to consistent physical states (e.g. via diffusion models), which we leave for future work.

\newpage
\bibliography{bib}
\bibliographystyle{icml2024}

%%%%%%%%%%%%%%%%%%%%%%%%%%%%%%%%%%%%%%%%%%%%%%%%%%%%%%%%%%%%%%%%%%%%%%%%%%%%%%%
%%%%%%%%%%%%%%%%%%%%%%%%%%%%%%%%%%%%%%%%%%%%%%%%%%%%%%%%%%%%%%%%%%%%%%%%%%%%%%%
% APPENDIX
%%%%%%%%%%%%%%%%%%%%%%%%%%%%%%%%%%%%%%%%%%%%%%%%%%%%%%%%%%%%%%%%%%%%%%%%%%%%%%%
%%%%%%%%%%%%%%%%%%%%%%%%%%%%%%%%%%%%%%%%%%%%%%%%%%%%%%%%%%%%%%%%%%%%%%%%%%%%%%%
\newpage
\appendix
\onecolumn

\section{Additional details}
\subsection{Training details}\label{app:training-details}

We denote $X_t$ the historical trajectory of ERA5, indexed by time $t$.
Input states $X_t$ are normalized to zero mean and unit variance on a per-variable and per-level basis, using statistics of the training set 1979-2018. We train the model to predict the difference $X_{t+\delta t} - X_t$, which we similarly normalize to unit variance.

Following GraphCast, we scale the training loss with coefficients proportional to the air density, to give more importance to variables closer to the surface. We also use the same reweighting of the surface variables with a coefficient of $1$ for 2m temperature, and $0.1$ for wind components and mean surface pressure.

We train our models for $300k$ steps with the AdamW optimizer \cite{kingma2014adam}. The batch size is $4$ and the optimizer parameters are a learning rate of 3e-4, beta parameters $(\beta_1=0.9, \beta_2=0.98)$ and a weight decay of $0.05$. The learning rate is increased linearly for the first 5000 steps, then decayed with a cosine schedule for the remaining steps.

\subsection{Comparison with state-of-the-art} \label{app:sota}

For all models except Stormer, RMSE scores at 1.5º are taken from WeatherBench2 \cite{rasp2023weatherbench}. For Stormer \cite{nguyen2023scaling}, we evaluate outputs provided by the authors at 1.4º resolution. Stormer is a $\sim$300M parameters model trained to forecast ERA5 variables at multiple lead-time simultaneously: 6h, 12h and 24h. To make a 24h lead-time forecast, Stormer uses all possible combinations of lead times as conditioning: 24h, 12h-12h, 12h-6h-6h, 6h-12h-6h, 6h-6h-12h, 6h-6h-6h-6h, and averages all trajectories. This base model is ran 16 times with different lead time conditioning to make a 24h forecast. Our ensemble models requires only two forward passes with $\sim$164M parameters, or four passes with $\sim$89M parameters. Please see the paper \cite{nguyen2023scaling} for more details on Stormer.

\subsection{Metrics}\label{app:metrics}

We compute the average RMSE improvement over the IFS HRES as 
$$\text{RRH}(model) = \frac{1}{\sum_v \alpha_v} \sum_v \alpha_v \frac{\text{RMSE}_{v}(HRES) - \text{RMSE}_{v}(model)}{\text{RMSE}_{v}(HRES)}$$ 
where variables $v$ spans a set $\mathcal{V}$ of representative weather variables: Z500, Q700, T850, U850, V850, T2m, SP, U10m, V10m. $\alpha_v$ is a per-variable scaling, which is $0.5$ for $U$ and $V$ and $1$ for all other variables. We use this scaling for wind variables instead of combining $U$ and $V$ predictions in a single wind vector score as in WeatherBench.

As usual, the RMSE scores of the IFS HRES are computed against the IFS analysis \cite{rasp2023weatherbench}.

\subsection{Convolutional Head}
\label{app:convhead}

\begin{figure}[h]
    \centering
    \includegraphics[width=0.7\linewidth]{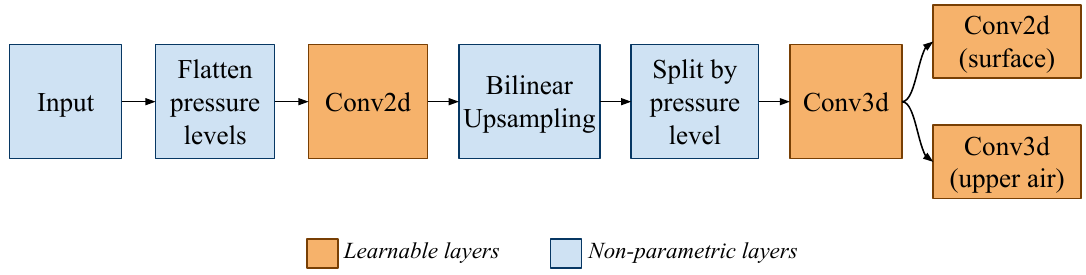}
    \caption{Architecture of our convolutional head.}
    
    \label{fig:convhead_design}
\end{figure}

\begin{figure}[ht]
    \centering
    \includegraphics[width=\linewidth]{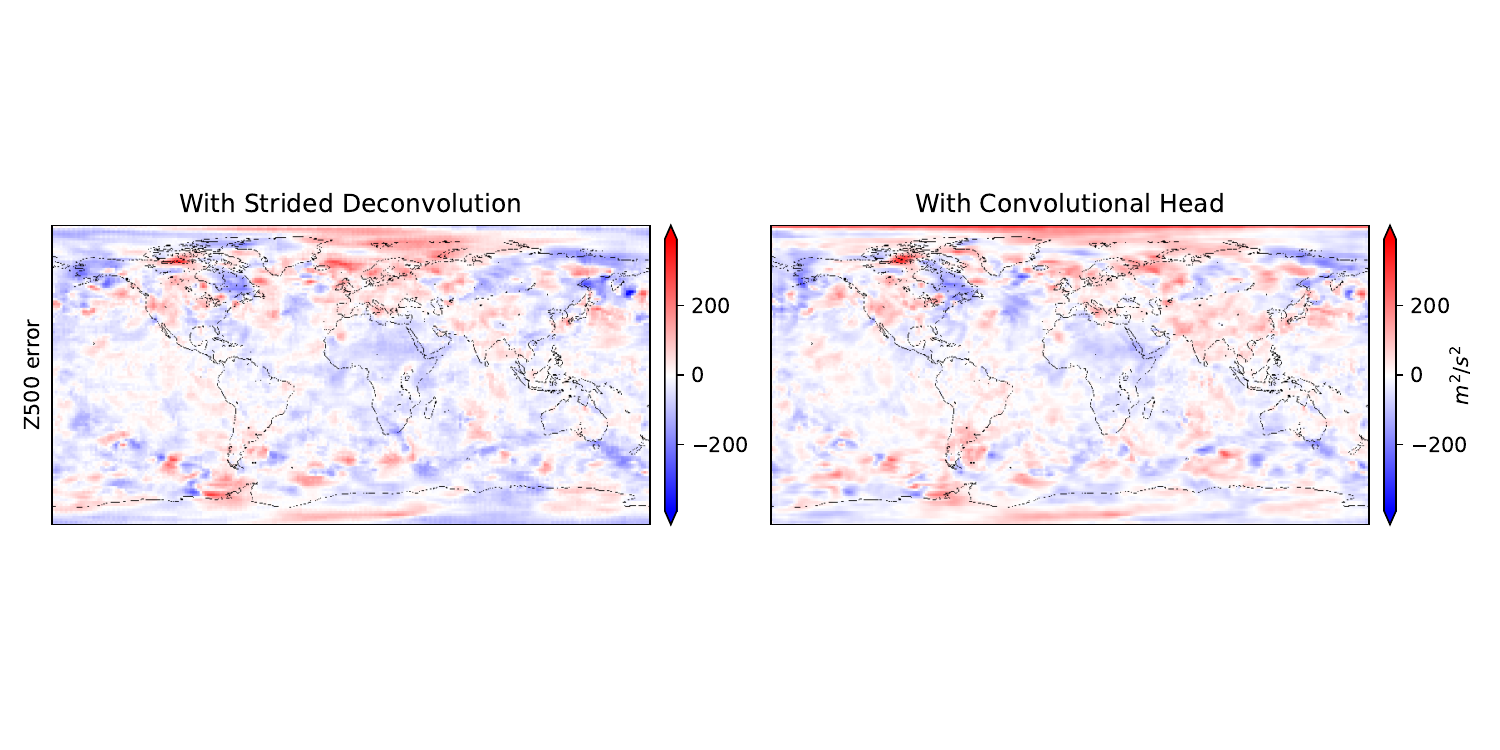}
    \caption{Z500 error using the transformer with strided deconvolution (left) versus the convolutional head that we use (right).}
    \label{fig:convhead}
\end{figure}

In early experiments, we have observed that the transformer architecture with a strided deconvolution produces small but noticeable checkerboard artefacts (see Figure \ref{fig:convhead}, notably near the North and South poles). Since these artefacts can cause problems for downstream applications, we design a convolutional head that smoothly upsamples data to recover the original image resolution instead. Our design is based on bilinear upsampling, and since it has no learnable parameters, we add convolutions before and after, see Figure \ref{fig:convhead_design}.

\section{Additional Results}

\subsection{Quantitative results}

In Table \ref{full_results}, we compare our model against all models available in WeatherBench2, including those trained at 0.25º resolution. In Table 4, we report metrics for all key weather variables in our ablation study, where we compare our final model with a version without finetuning on recent data (w/o FT) and a version without finetuning and without our proposed Cross-Level-Attention (w/o CLA). We also report scores for small ensembles of our models. Due to computational constraints, we do not train versions w/o CLA for the large (L) model and only train L versions for the ensemble.

\input{figs/full_results}

\input{figs/full_ablation_table}

\subsection{Multi-step evaluation}\label{app:multistep}

In the main paper, we only evaluated models with a lead time of 24h. In this section, we evaluate models for longer lead times through auto-regressive rollouts. After the 300k steps of training, models are fine-tuned with 20k of multi-step fine-tuning by rolling out the model $K$ times and averaging losses at each step. We use $K=2$ for the first 8k steps, $K=3$ for the next 8k steps and $K=4$ for the remaining 4k steps \cite{keisler2022forecasting, nguyen2023scaling, lam2022graphcast}.
Interestingly, we find that our model \sct{ArchesWeather-M} performs better than the original Pangu-Weather and even GraphCast at longer lead times for all variables, which might be due to a smoothing effect due to the corser resolution. The NeuralGCM ensemble still performs better for upper-air variables (surface variables are not predicted by the model), since the ensemble mean with 50 members better approximate the true distribution mean, but we can partly close this gap and match the performance of Stormer with our \sct{ArchesWeather-Mx4} ensemble version. We also note that the better 24h RMSE scores for Stormer on wind variables (U850, V850, U10m, V10m) do not yield better multi-step trajectories as the week-ahead wind predictions of our ensemble model are competitive with and even slightly outperform Stormer.

\begin{figure}[ht]
    \centering
    \includegraphics[width=\linewidth]{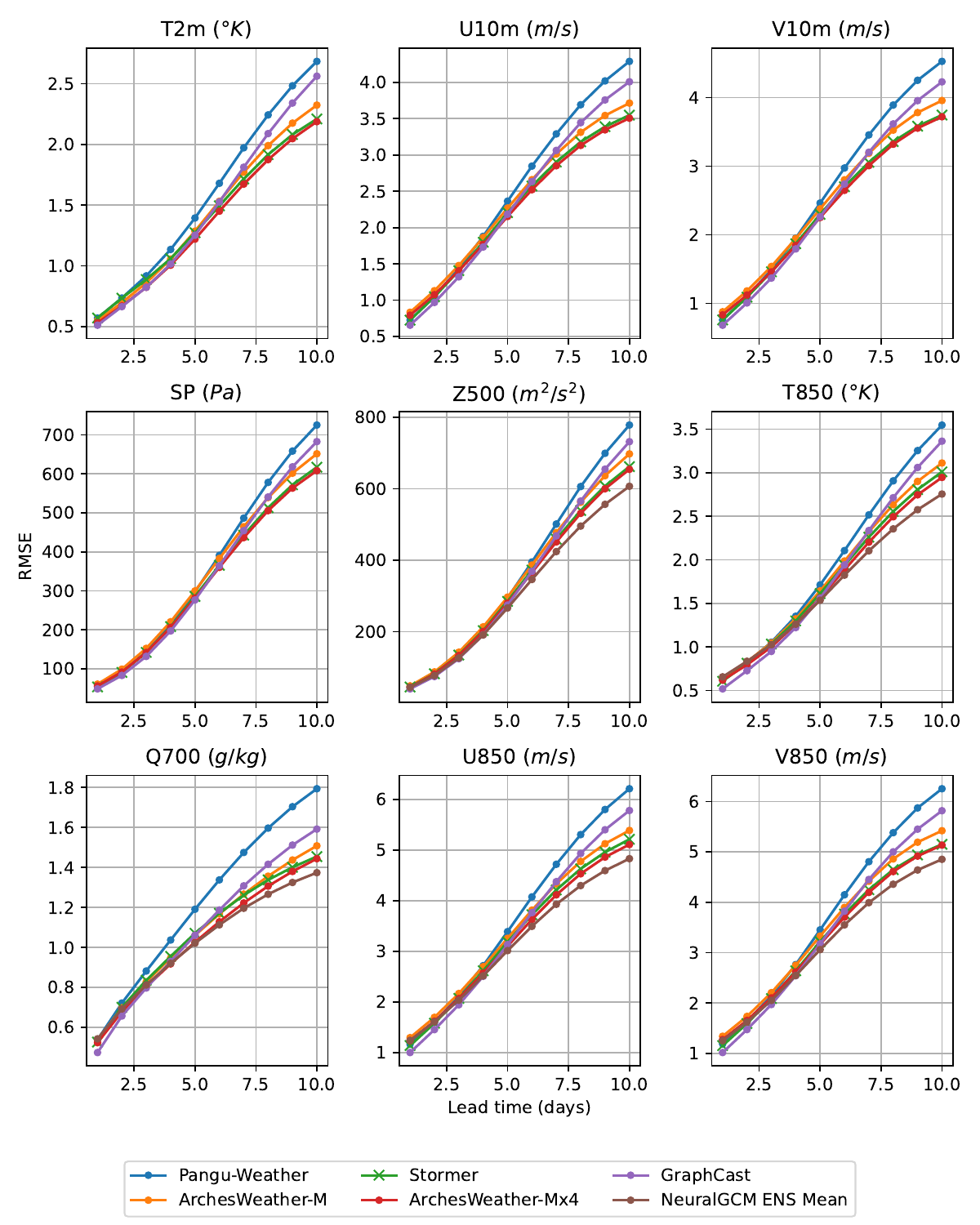}
    \caption{RMSE scores of weather models for lead times up to 10 days.}
    \label{fig:multistep}
\end{figure}

\subsection{Qualitative results}

\begin{figure}[p]
    \centering
    \includegraphics[width=\linewidth]{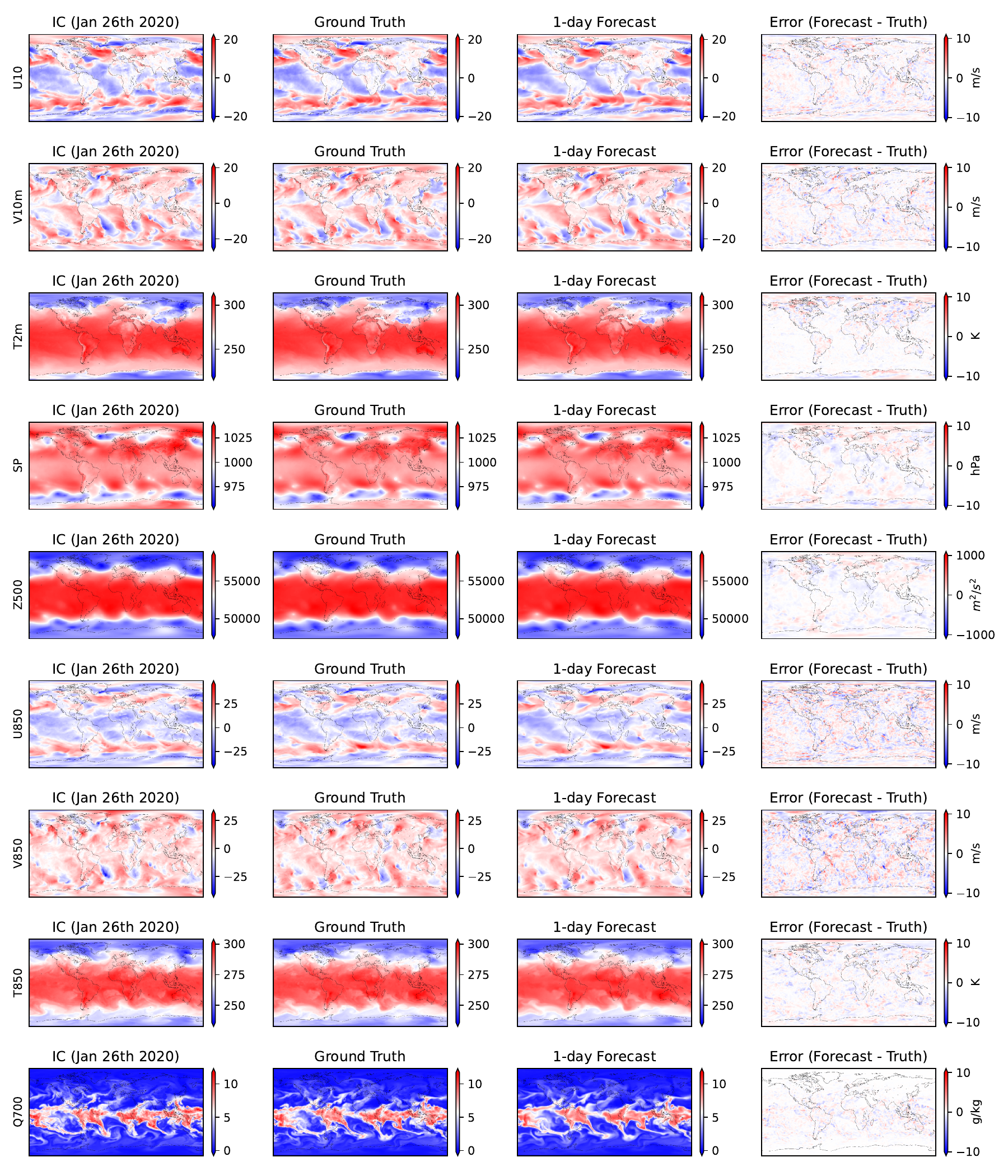}
    \caption{\sct{ArchesWeather-M} forecasts, initialized the 26th of January 2020.}
    \label{fig:forecast-1d-raw}
\end{figure}

\begin{figure}[p]
    \centering
    \includegraphics[width=\linewidth]{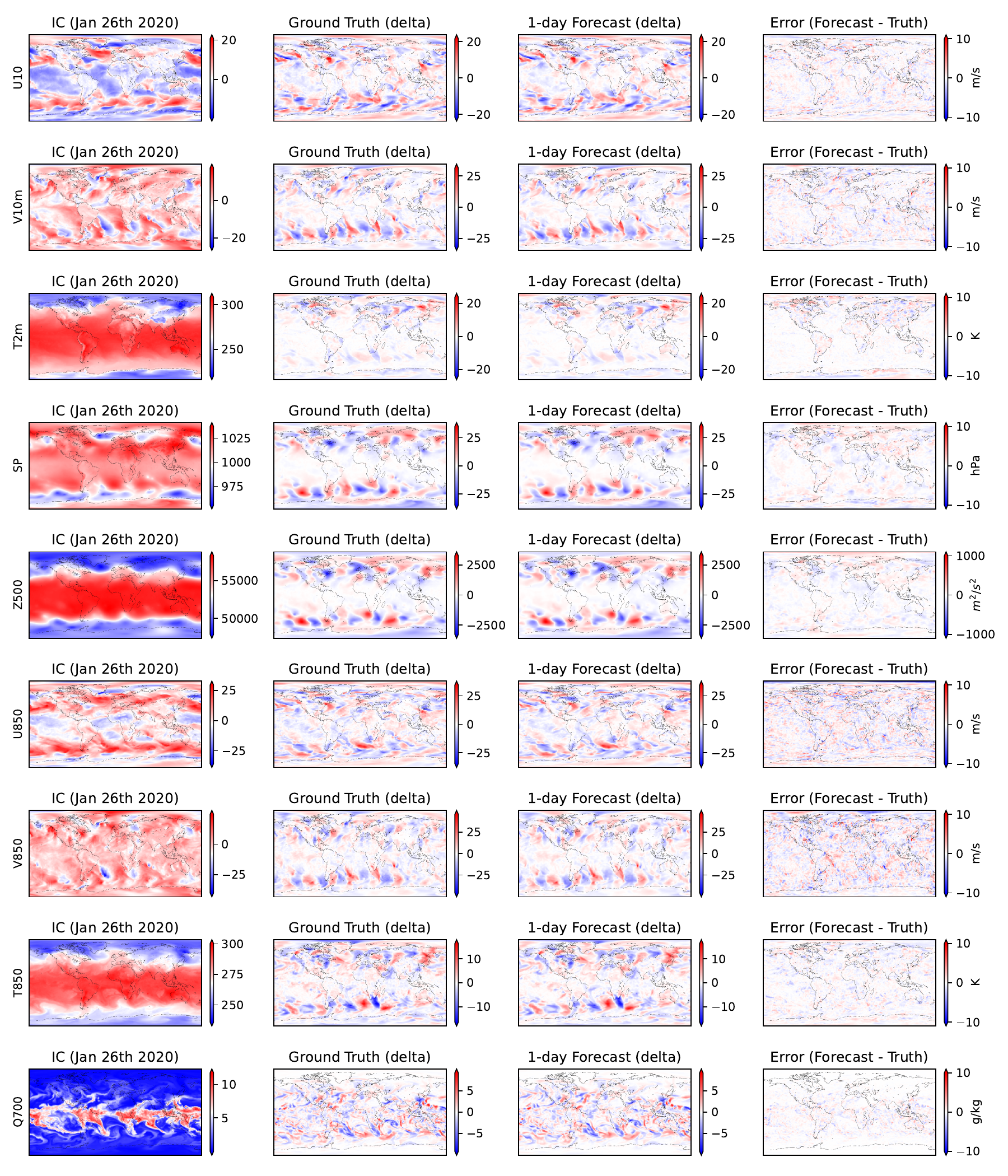}
    \caption{\sct{ArchesWeather-M} forecasts, initialized the 26th of January 2020. The state increments ($X_{t+\delta t} - X_t$) are shown.}
    \label{fig:forecast-1d-delta}
\end{figure}

Qualitative samples for \sct{ArchesWeather-M} are shown in Figure \ref{fig:forecast-1d-raw} (raw forecasts $X_t$) and \ref{fig:forecast-1d-delta} (predicted deltas $X_{t+\delta t} - X_t$). We chose January 26th as initialization date, similarly to the qualitative results in Stormer \cite{nguyen2023scaling}.

%% file: figs/main_table.tex
\begin{table*}[t]
  \centering
\vskip 0.15in
\begin{small}
\begin{sc}
\begin{tabular}{lllccccccccc}
\toprule
                  & \textbf{Res.} & \textbf{Cost} & \textbf{Z500} & \textbf{T850} & \textbf{Q700} & \textbf{U850} & \textbf{V850} & \textbf{T2m} & \textbf{SP} & \textbf{U10m} & \textbf{V10m} \\
\midrule
IFS HRES              &    0.1º           &                                                                & 42.30         & 0.625         & 0.556         & 1.186         & 1.206         & 0.513        & 60.16       & 0.833        & 0.872        \\
\midrule
Pangu             & 0.25º         & 2880                                                           & 44.31         & 0.620         & 0.538         & 1.166         & 1.191         & 0.570        & 55.14       & 0.728        & 0.759        \\
GraphCast         & 0.25º         & 2688                                                           & 39.78         & 0.519         & 0.474         & 1.000         & 1.02          & 0.511        & 48.72       & 0.655        & 0.683        \\
\midrule
%Keisler GNN           & 1º            & 11                                                             & 66.87         & 0.816         & 0.658         & 1.584         & 1.626         & N/A          & N/A         & N/A          & N/A          \\
SphericalCNN      & 1.4º          & 384                                                            & 54.43         & 0.738         & 0.591         & 1.439         & 1.471         & N/A          & N/A         & N/A          & N/A          \\
Stormer           & 1.4º          & 256                                                            & 45.12         & \bfu{0.607}         & 0.527         & \bfu{1.138}         & \bfu{1.156}         & 0.570        & \bfu{53.77}       & \bfu{0.726}        & \bfu{0.760}        \\
NeuralGCM ENS (50) & 1.4º          & 7680                                                           & \bfu{43.99}         & 0.658         & 0.540         & 1.239         & 1.256         & N/A          & N/A         & N/A          & N/A          \\
\midrule
ArchesWeather-M           & 1.5º          & 11                                                            & 48.1          & 0.645         & 0.538         & 1.294         & 1.342         & 0.550        & 60.9        & 0.834        & 0.877        \\
ArchesWeather-L & 1.5º & 22 & 46.32 & 0.621 & 0.530 & 1.242 & 1.286 & 0.540 & 58.649 & 0.798 & 0.838 \\

ArchesWeather-Mx4 & 1.5º          & 44                                                           & \textbf{44.36} & \textbf{0.619} & \textbf{0.523}	&1.235	&1.277	& 0.530	& 56.3	& 0.793	& 0.832       \\
ArchesWeather-Lx2 & 1.5º & 44 & \textbf{44.35} & \bfu{0.606} & \bfu{0.519} & \textbf{1.207} & \textbf{1.251} & \bfu{0.525} & \textbf{55.956} & \textbf{0.776} & \textbf{0.815} \\
\bottomrule

\end{tabular}
\end{sc}
\end{small}
\vskip -0.05in
  \caption{Comparison of AI weather models on RMSE scores for key weather variables with 24h lead-time. Cost is the training computational budget in V100-days. Best scores for training resolution coarser than 1º in \bfu{underlined bold}, second best scores in \textbf{bold}.}
  \label{table:main}
  \vskip -1.3em
\end{table*}

%48.02	0.6435	0.5399	1.2905	1.3362	0.5514	60.9819	0.8296	0.8723

%% file: figs/ablation_table.tex
\begin{table}[ht]
  \centering
\vskip 0.15in
\begin{small}
\begin{sc}
\begin{tabular}{r|lc|ccc}
\toprule
 & \textbf{Model} & \#L & \textbf{Z500} & \textbf{T2m} & \textbf{RRH↑} \\
\midrule
A1 & Pangu-S & 16 & 66.7 & 0.84 & -30.6 \\
B1 & ArchesWeather-S & 16 & 49.3 & 0.566 & -8.6 \\
C1 & \; - w/o fine-tuning & 16 & 50.6 & 0.567 & -9.7 \\
D1 & \; - w/o CLA\hspace{-0em} & 16 & 55.1 & 0.594 & -17.1 \\

\midrule
A2 & Pangu-M & 32 & 58.7 & 0.78 & -20.4  \\
B2 & ArchesWeather-M & 32 & 48.0 & 0.551 & -5.0 \\
C2 & \; - w/o fine-tuning & 32 & 48.7 & 0.552 & -5.6 \\
D2 & \; - w/o CLA\hspace{-0em}  & 32 & 51.8 & 0.572 & -11.6 \\

\bottomrule
\end{tabular}
\end{sc}
\end{small}
\vskip -0.1em
\caption{500hPa geopotential and 2m temperature RMSE at 24h lead-time for different version of our models, and Pangu-Weather re-trained at 1.5º. \sct{ArchesWeather w/o CLA} uses local 3D attention instead of our proposed Cross-Level Attention. RRH is the relative RMSE improvement over HRES.}
\label{abl}
\vskip -1.7em
\end{table}

%% file: figs/full_results.tex
\begin{table}[ht]
  \centering
\begin{small}
\begin{sc}
\begin{tabular}{lllccccccccc}
\toprule
                  & \textbf{Res.} & \textbf{Cost} & \textbf{Z500} & \textbf{T850} & \textbf{Q700} & \textbf{U850} & \textbf{V850} & \textbf{T2m} & \textbf{SP} & \textbf{U10m} & \textbf{V10m} \\
\midrule
IFS               &               &                                                                & 42.30         & 0.625         & 0.556         & 1.186         & 1.206         & 0.513        & 60.16       & 0.833        & 0.872        \\
\midrule
Pangu-Weather             & 0.25º         & 2880                                                           & 44.31         & 0.620         & 0.538         & 1.166         & 1.191         & 0.570        & 55.14       & 0.728        & 0.759        \\
NeuralGCM         & 0.25º         & 16128                                                          & 37.94         & 0.547         & 0.488         & 1.050         & 1.071         & N/A          & N/A         & N/A          & N/A          \\
FuXi              & 0.25º         & 52                                                             & 40.08         & 0.548         & N/A           & 1.034         & 1.055         & 0.532        & 49.23       & 0.660        & 0.688        \\
GraphCast         & 0.25º         & 2688                                                           & 39.78         & 0.519         & 0.474         & 1.000         & 1.02          & 0.511        & 48.72       & 0.655        & 0.683        \\
\midrule
Keisler           & 1º            & 11                                                             & 66.87         & 0.816         & 0.658         & 1.584         & 1.626         & N/A          & N/A         & N/A          & N/A          \\
SphericalCNN      & 1.4º          & 384                                                            & 54.43         & 0.738         & 0.591         & 1.439         & 1.471         & N/A          & N/A         & N/A          & N/A          \\
Stormer           & 1.4º          & 256                                                            & 45.12         & 0.607         & 0.527         & 1.138         & 1.156         & 0.570        & 53.77       & 0.726        & 0.760        \\
NeuralGCMEns (50) & 1.4º          & 7680                                                           & 43.99         & 0.658         & 0.540         & 1.239         & 1.256         & N/A          & N/A         & N/A          & N/A          \\
\midrule
ArchesWeather-M           & 1.5º          & 9                                                            & 48.0          & 0.643         & 0.539         & 1.290         & 1.336         & 0.551        & 60.9        & 0.829        & 0.872        \\
ArchesWeather-L & 1.5º & 18 & 46.32 & 0.621 & 0.530 & 1.242 & 1.286 & 0.540 & 58.649 & 0.798 & 0.838 \\

ArchesWeather-M4 & 1.5º          & 36                                                           & 43.91 & 0.616 & 0.522	&1.230	&1.271	&0.528	& 55.69	&0.789	&0.828       \\
ArchesWeather-L2 & 1.5º & 36 & 44.35 & 0.606 & 0.519 & 1.207 & 1.251 & 0.525 & 55.956 & 0.776 & 0.815 \\
\bottomrule

\end{tabular}
\end{sc}
\end{small}
\vskip -0.1in
  \caption{RMSE scores for \ours~ compared to all weather forecasting models available in WeatherBench2.}
  \label{full_results}
\end{table}

%% file: figs/full_ablation_table.tex
\begin{table}[ht]
\label{full_ablation_tab}
  \centering
\vskip 0.15in
\begin{small}
\begin{sc}
\begin{tabular}{lllccccccccc}
\toprule
L & Name & Z500 & T850 & Q700 & U850 & V850 & T2m & SP & U10m & V10m \\
\midrule
2 & ArchesWeather-S w/o CLA & 55.116 & 0.722 & 0.577 & 1.432 & 1.485 & 0.594 & 69.768 & 0.942 & 0.993 \\
29 & ArchesWeather-S w/o FT & 50.632 & 0.676 & 0.554 & 1.354 & 1.402 & 0.567 & 63.447 & 0.878 & 0.922 \\
9 & ArchesWeather-S & 49.365 & 0.672 & 0.551 & 1.345 & 1.395 & 0.567 & 62.710 & 0.870 & 0.914 \\
6 & ArchesWeather-Sx4 & 47.042 & 0.652 & 0.541 & 1.299 & 1.347 & 0.549 & 59.291 & 0.838 & 0.881 \\
\midrule
10 & ArchesWeather-M w/o CLA & 51.820 & 0.688 & 0.558 & 1.393 & 1.425 & 0.572 & 65.616 & 0.888 & 0.934 \\
25 & ArchesWeather-M w/o FT & 48.720 & 0.646 & 0.541 & 1.295 & 1.340 & 0.552 & 61.572 & 0.834 & 0.877 \\
18 & ArchesWeather-M & 48.022 & 0.644 & 0.540 & 1.290 & 1.336 & 0.551 & 60.981 & 0.830 & 0.872 \\
39 & ArchesWeather-Mx4 & 43.911 & 0.616 & 0.522 & 1.230 & 1.271 & 0.528 & 55.688 & 0.789 & 0.828 \\
\midrule
23 & ArchesWeather-L w/o FT & 46.846 & 0.623 & 0.531 & 1.245 & 1.288 & 0.540 & 59.031 & 0.801 & 0.841 \\
44 & ArchesWeather-L & 46.321 & 0.621 & 0.530 & 1.242 & 1.286 & 0.540 & 58.649 & 0.798 & 0.838 \\
18 & ArchesWeather-Lx2 & 44.346 & 0.606 & 0.519 & 1.207 & 1.251 & 0.525 & 55.956 & 0.776 & 0.815 \\
\bottomrule
\end{tabular}
\end{sc}
\caption{Ablation: RMSE scores of key weather variables for variants of our model. w/o FT does not use recent data for the last 50k steps, and w/o CLA additionally does not use Cross-Level-Attention. The -Sx4, -Mx4 and -Lx2 are models ensembles using 4, 4, and 2 models respectively.}
\end{small}

\end{table}